\documentclass[letterpaper, 10pt, conference]{ieeeconf}      

\IEEEoverridecommandlockouts                              

\usepackage{enumerate}
\usepackage{tikz}
\usetikzlibrary{shapes.geometric, arrows}
\usetikzlibrary{shapes,arrows}
\usepackage{cite}
\usepackage{float}
\usepackage{xfrac}
\usepackage{graphics} 
\usepackage{tabularx} 
\usepackage{graphicx}
\usepackage{epstopdf}
\usepackage{epsfig} 
\usepackage{mathptmx} 
\usepackage{times} 
\usepackage{amsmath} 
\usepackage{amssymb}  
\usepackage{caption}
\usepackage{mathtools}
\usepackage{comment}
\usepackage{graphicx}
\usepackage{subfig}
\usepackage{soul}
\usepackage{caption}
\usepackage{subcaption}
\usepackage{multirow}

\usepackage{algorithm}
\usepackage{algpseudocode}%

\definecolor{mygreen}{rgb}{0.10,0.50,0.10}

\newtheorem{theorem}{Theorem}

\newtheorem{corollary}[theorem]{Corollary}

\DeclarePairedDelimiter\abs{\lvert}{\rvert}
\makeatletter
\let\oldabs\abs
\def\abs{\@ifstar{\oldabs}{\oldabs*}}
\newcommand{\RN}[1]{%
  \textup{\uppercase\expandafter{\romannumeral#1}}%
}
\DeclareMathOperator*{\argmin}{arg\,min}

\title{\LARGE \bf
Autonomous helicopter aerial refueling: controller design and performance guarantees
}

\author{Damsara Jayarathne$^{1}$ Santiago Paternain$^{2}$ and Sandipan Mishra$^{1}$
\thanks{$^{1}$Damsara Jayarathne and Sandipan Mishra are with the Department of Mechanical, Aerospace and Nuclear Engineering, while
$^{2}$Santiago Paternain is with the Department of Electrical, Computer, and Systems Engineering at the 
Rensselaer Polytechnic Institute in Troy, NY.
        {\tt\small \{jayarj2,paters,mishrs2\}@rpi.edu}}%
}

\begin{document}

\maketitle
\thispagestyle{empty}
\pagestyle{empty}

\begin{abstract}
 
In this paper, we present a control design methodology, stability criteria and performance bounds for autonomous helicopter aerial refueling. Autonomous aerial refueling is particularly difficult due to the aerodynamic interaction between the wake of the tanker, the contact-sensitive nature of the maneuver, and the uncertainty in drogue motion. Since the probe tip is located significantly away from the helicopter's center-of-gravity, its position (and velocity) is strongly sensitive to the helicopter's attitude (and angular rates). In addition, the fact that the helicopter is operating at high speeds to match the velocity of the tanker forces it to maintain a particular orientation, making the docking maneuver especially challenging. In this paper, we propose a novel outer-loop position controller that incorporates the probe position and velocity into the feedback loop. The position and velocity of the probe tip depend both on the position (velocity) and on the attitude (angular rates) of the aircraft. We derive analytical guarantees for docking performance in terms of the uncertainty of the drogue motion and the angular acceleration of the helicopter, using the ultimate boundedness property of the closed-loop error dynamics. Simulations are performed on a high-fidelity UH60 helicopter model with a high-fidelity drogue motion under wind effects to validate the proposed approach for realistic refueling scenarios. These high-fidelity simulations reveal that the proposed control methodology yields an improvement of 36\% in the 2-norm docking error compared to the existing standard controller.

\end{abstract}
\section{Introduction}


Helicopter aerial refueling is an aerial operation where a helicopter is refueled in mid-flight from a fixed-wing tanker aircraft. The goal of this maneuver is to guide the probe of the helicopter to dock onto the drogue which is attached to and trailing from the tanker aircraft. Helicopter aerial refueling is a particularly challenging maneuver because of the 1) aerodynamic interaction between the tanker, the drogue, and the helicopter, 2) strict safety constraints, and 3) limited time to dock. Therefore, there is a need to develop autonomous control algorithms for reducing pilot workload and improving safety of these maneuvers.

Autonomous aerial refueling for \textit{fixed-wing} aircrafts has been extensively studied \cite{mao2008survey, tandale2006trajectory, martinez2013vision} and has seen significant success in the recent past. Recently, there has been an increase in interest in autonomous helicopter aerial refueling \cite{jayarathne2023safe, jayarathne2024allocation} mainly due to the advancements in tools to analyze complex aerodynamic interactions \cite{loechert2021consideration} and knowledge of high-fidelity simulation platforms. For most autonomous maneuvers, the typical helicopter control architecture consists of an inner-loop controller that controls the attitude and an outer-loop controller that controls the motion in the longitudinal and lateral directions \cite{zhao2021differential, horn2015autonomous, Krishnamurthi_2017_Flight}.
Autonomous helicopter control strategies have been developed for shipboard-landing \cite{zhao2021differential, horn2015autonomous} using dynamic inversion control, formation flying \cite{karimoddini2013hybrid} using hybrid supervisory control, and slung load carrying \cite{geng2020cooperative, enciu2017flight} using a cascaded (linear) controller. 

Specifically, contact-based maneuvers such as shipboard landing \cite{saripalli2006landing, zhao2021differential} and slung load carrying require having control over points that lie outside the center of gravity (CG) of the helicopter while controlling the attitudes carefully. In the case of aerial helicopter refueling, the probe tip lies several feet forward of the helicopter CG, making it extremely sensitive to the attitude and angular rates of the helicopter. Furthermore, the helicopter is flying at approximately 110 knots ($56.58 m/s$), close to its maximum speed, reducing the available control authority during the maneuver. In our previous work \cite{jayarathne2023safe, jayarathne2024allocation}, we designed a controller for the helicopter aerial refueling maneuver that used drogue position/velocity and helicopter CG position/velocity as the feedback error signal to drive the outer-loop. However, this control law did not consider the probe orientation explicitly, and therefore potentially could result in a higher error at the time of docking, especially with significant drogue motion uncertainty.


One possible approach for contact-based maneuvers is to use iterative, receding-horizon frameworks such as robust Model Predictive Control (MPC) \cite{Ngo_2016_Model,chung2006autonomous}, which account for model and parameter uncertainty, and strict safety constraints \cite{misra2018stochastic}. However, MPC algorithms that utilize full-state nonlinear models require significant computational power, which make them less suitable for real-time implementation. Moreover, it is challenging to obtain analytical guarantees on performance and robustness for these iterative receding-horizon frameworks. Finally, in the context of helicopter aerial refueling, the complexity of the drogue motion model can also make MPC algorithm design challenging, given the natural uncertainty of the drogue motion.

On the other hand, analytical model-based control strategies can provide explicit stability certificates along with quantification of performance \cite{halbe2020robust} and robustness to modeling uncertainty. This paper proposes a new nonlinear dynamic-inversion-based outer-loop position control law that explicitly takes the probe orientation into account. The proposed controller adjusts the commanded acceleration to the helicopter according to the probe position and velocity, which is a function of the angular and positional states of the helicopter.  The main contributions of this paper are four-fold, 1) designing a controller that takes the orientation of the probe into account 2) proving the stability of the proposed controller, 3) providing a theoretical guarantee for docking performance under drogue and helicopter motion uncertainty, and 4) validating the proposed controller on a UH60 Black Hawk
simulator with realistic drogue dynamics and wind effects. 

\section{Preliminaries}

\subsection{Helicopter aerial refueling}

\begin{figure}[h!]
\captionsetup{font=footnotesize}
\centering
\includegraphics[width=0.48\textwidth]{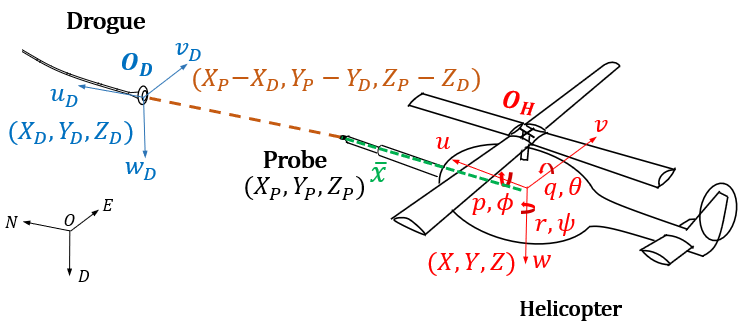}
\caption{Schematic of helicopter aerial refueling. $O$, $O_H$, $O_D$ correspond to the North-East-Down, helicopter's and drogue's coordinate frames, respectively. The velocity components, the rates of rotation, and the Euler angles in the $O_H$ coordinate frame are ($u,v,w$), ($p,q,r$) and ($\phi,\theta,\psi$), respectively. The velocity of the drogue is $u_D,v_D,w_D$. ($X,Y,Z$), ($X_P,Y_P,Z_P$), and ($X_D,Y_D,Z_D$) are the positions of the helicopter, the probe and the drogue, respectively. $\bar{x}$ is the vector that defines the probe tip in $O_H$.}
\label{Problem_formulation}
\end{figure}

Figure \ref{Problem_formulation} illustrates the general helicopter aerial refueling problem. At the beginning of the maneuver at time $t_0$, the helicopter fuselage and the drogue are at states $\mathbf{x}_{f0}$ and $\mathbf{x}_{D0}$, respectively. 
The goal is to guide the helicopter probe to dock with the drogue at time $T$, by matching their states to within a tolerance $\epsilon_0$.
The state of the drogue is described by $\textbf{x}_{D}(\cdot)$, which is unpredictable due to the air wake of the tanker and the so-called \textit{bow-wave} effect. We formulate the autonomous aerial refueling problem as follows. Find $\{\mathbf{x}^*(\cdot), T, \mathbf{u}^*(\cdot) \}$ as
\begin{eqnarray}\label{eqn_ConstraintsHAAR}
&\begin{array}{c} 
 \argmin J(\mathbf{x}(\cdot), T, \mathbf{u}(\cdot))\\
\dot{\mathbf{x}}(t) = f(\mathbf{x}(t), \mathbf{u}(t)) \qquad \textrm{Helicopter dynamics} \\
\mathbf{x}_{f}(0) = \mathbf{x}_{f0} \qquad \textrm{Initial helicopter state} \\
\dot{\mathbf{x}}_{p}(t) =  G(\mathbf{x}_{f}(t), \bar{x}) \qquad
\textrm{Probe state trajectory} \\
|| \mathbf{x}_{D}(T) - \mathbf{x}_{p}(T)|| \leq \epsilon_0 \qquad
\textrm{Docking criterion} 
\end{array}
\end{eqnarray}
 where $t$, $\mathbf{x}_{f}$, $\mathbf{x}_{p}$ and $\mathbf{x}_{D}$ are time, the fuselage states, states of the probe and the drogue, respectively. $\bar{x}$ is the relative distance between the center of gravity of the helicopter and the probe tip in helicopter coordinate frame $O_H$. $G$ is the mapping from the helicopter fuselage center of gravity to the probe tip $\textbf{x}_p$. Note that for minimizing docking time,  $J(\mathbf{x}(\cdot), T, \mathbf{u}(\cdot)) = T$. 

\subsection{General two-loop helicopter control architecture} \label{sec_HCcontrol}

\begin{figure}[h!]
\captionsetup{font=footnotesize}
\centering
\vspace{2mm}
\includegraphics[width=0.48\textwidth]{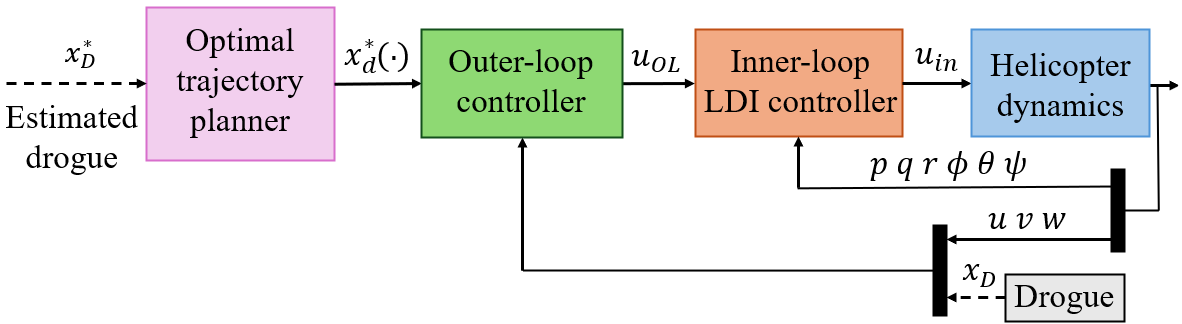}
\caption{Generic inner-loop outer-loop control architecture.}
\label{Fig_GenControl}
\end{figure}

The general helicopter control architecture consists of an inner-loop and an outer-loop \cite{zhao2021differential, geng2020cooperative, chung2006autonomous} as shown in Figure \ref{Fig_GenControl}. The inner-loop utilizes the well-known (scheduled) Linear Dynamic Inversion (LDI) method to control the roll $\phi$, pitch $\theta$, vertical velocity $\dot{Z}$, and yaw rate $\dot{\psi}$ of the helicopter. It is key to note that this LDI controller does not regulate the lateral and longitudinal dynamics of the helicopter, these so-called \textit{zero-dynamics} are regulated using the outer-loop controller, which takes position and velocities in the longitudinal and lateral directions into account to generate appropriate attitude (and altitude) commands for the inner loop. The outer-loop controller which generates the outer-loop control commands $u_{OL} = [{\phi}_c,\theta_c,\ddot{Z}_c,\dot{\psi}_c]^T$ utilizes a dynamic inversion controller and a PD controller. $\textbf{x}_d^*(\cdot)$ represents the reference trajectories generated by the trajectory planner \cite{zhao2021differential}. These trajectories include the closure between the drogue and the helicopter $x^*-x_D^*$. The control commands $u_{in} = [u_{lat}, u_{lon}, u_{col}, u_{ped}]^T$ generated by the outer-loop which includes the lateral, longitudinal, collective, and pedal controls are sent to the helicopter dynamics.

\begin{figure*}[t] 
\captionsetup{font=footnotesize}
\centering
\vspace{2mm}
\includegraphics[width=1\textwidth]{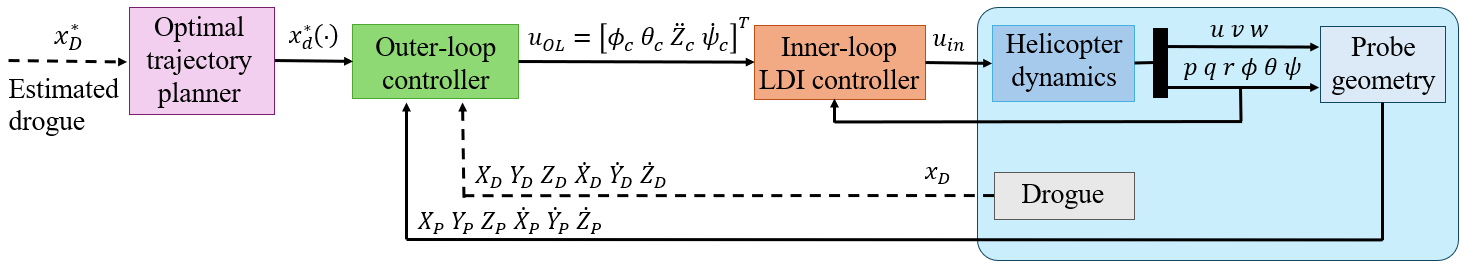}
\caption{Proposed control architecture for helicopter aerial refueling.}
\label{HCControlArch} 
\end{figure*}

The proposed controller consists of an inner-loop and outer-loop as illustrated in  Figure \ref{HCControlArch}. 
The probe geometry which includes the probe position and velocity is calculated using the states $u, v, w, p, q, r, \phi, \theta, \psi$ which represent the velocities, angular rates, and attitudes of the aircraft. The probe position and velocity are fed to the outer loop to generate the commanded acceleration of the CG of the helicopter.   

\section{Mathematical modelling}


In this section, first, we formulate the outer-loop helicopter dynamics (Section \ref{sec_HCDynamics}). 
We then derive the error dynamics between the probe and the drogue.

\subsection{Helicopter dynamics for outer-loop control design}
\label{sec_HCDynamics}

\noindent The outer-loop helicopter dynamics model is given by~\cite{zhao2021differential} \begin{eqnarray}\label{eqn_OLDynamics}
&\begin{array}{ccccc}
\ddot{X} = -g\left(\tan(\theta-\bar{\theta})\cos\psi+\cfrac{\tan(\phi-\bar{\phi})}{\cos(\theta-\bar{\theta})}\sin\psi\right) \\

\ddot{Y}= -g\left(\tan(\theta-\bar{\theta})\sin\psi-\cfrac{\tan(\phi-\bar{\phi})}{\cos(\theta-\bar{\theta})}\cos\psi\right). 
\end{array}
\end{eqnarray}

These dynamics assume instantaneous attitude and altitude-rate dynamics. This is reasonable given the time scale difference between the inner and outer loops. $X$ and $Y$ correspond to the position of the helicopter's CG in the North and East directions, respectively. 
Note that dynamics $\ddot{X}$ and $\ddot{Y}$ are the so-called \textit{zero dynamics} of the system. $\theta$, $\phi$, and $\psi$ correspond to the pitch, roll, and yaw angles, respectively. $\bar\theta$ and $\bar \phi$ refer to the trim attitudes.

\subsection{Standard outer-loop (position) controller}
\label{sec_StandardControl}

In this section, we discuss the typical control strategy for regulating the horizontal and vertical motion of the helicopter given by \cite{jayarathne2023safe}. The commanded accelerations (to the inner loop) are given by
\begin{align} \label{eq_CommAcc}
    \small
    \begin{bmatrix}
        \ddot{X}_{c} \\
        \ddot{Y}_{c} \\
        \ddot{Z}_{c}
    \end{bmatrix} = 
    \begin{bmatrix}
        \ddot{X}^* \\
        \ddot{Y}^* \\
        \ddot{Z}^* 
    \end{bmatrix} + 
    \begin{bmatrix}
        K_{dX} & 0 & 0\\
        0 & K_{dY} & 0 \\
        0 & 0 & K_{dZ} 
    \end{bmatrix} 
    \left(
    \begin{bmatrix}
        \dot{X}^* -  \dot{X}_{D}^* \\
        \dot{Y}^* -  \dot{Y}_{D}^* \\
        \dot{Z}^* -  \dot{Z}_{D}^* 
    \end{bmatrix} - 
    \begin{bmatrix}
        \dot{X} -  \dot{X}_{D} \\
        \dot{Y} -  \dot{Y}_{D} \\
        \dot{Z} -  \dot{Z}_{D}
    \end{bmatrix} 
    \right) \nonumber \\ 
    + 
    \begin{bmatrix}
        K_{pX} & 0 & 0\\
        0 & K_{pY} & 0 \\
        0 & 0 & K_{pZ} 
    \end{bmatrix}
        \left(
    \begin{bmatrix}
        X^* - X^*_{D} \\
        Y^* - Y^*_{D} \\
        Z^* - Z^*_{D}
    \end{bmatrix} - 
    \begin{bmatrix}
        X - X_{D} \\
        Y - Y_{D} \\
        Z - Z_{D}
    \end{bmatrix}  
    \right)
\end{align}

where $\ddot{X}_c$, $\ddot{Y}_c$, $\ddot{Z}_c$ are the \textit{commanded} accelerations in North, East and Down directions, respectively; $\ddot{X}^*$, $\ddot{Y}^*$, $\ddot{Z}^*$ are the desired accelerations; $K_{pX}$, $K_{pY}$, $K_{pZ}$,  $K_{dX}$, $K_{dY}$, $K_{dZ}$ are the PD gains of the controller;  $X$, $Y$, $Z$ $\dot{X}$, $\dot{Y}$, $\dot{Z}$ are the helicopter position and velocity in North-East-Down (NED) coordinate frame, $X_{D}$, $Y_{D}$,  $Z_{D}$, $\dot{X}_{D}$, $\dot{Y}_{D}$, $\dot{Z}_{D}$ are the drogue position and velocity in NED coordinate frame. Then, these commanded accelerations are used to generate the commanded pitch $\theta_c$ and roll $\phi_c$ to the inner loop. 

\textit{Remark}: Note that the commanded acceleration $\ddot{Z}_c$ is directly sent to the inner-loop to control the vertical motion.

By inverting the plant given in \eqref{eqn_OLDynamics}, we have
\begin{align}
\label{eqn_ControlInput}
\theta_{c} &= -\tan^{-1}\left( \dfrac{\ddot{X}_c}{g}\cos\psi_c + \dfrac{\ddot{Y}_c}{g}\sin\psi_c \right) + \bar{\theta} \nonumber \\
\phi_{c} &= \tan^{-1}\left(  \dfrac{-\ddot{X}_c\sin\psi_c + \ddot{Y}_c\cos\psi_c}{\sqrt{g^2 + (\ddot{X}_c\sin\psi_c + \ddot{Y}_c\cos\psi_c)^2 }} \right) + \bar{\phi} 
\end{align}

where $g$ is the gravitational acceleration; $\psi_c$ is the commanded yaw angle. Note that the commanded yaw rate is directly computed in the planning stage as 
\begin{align}
\label{eqn_ControlInput_Zpsi}
\dot{\psi}_{c} &= \dot{\psi}^*.
\end{align} 




\subsection{Proposed outer-loop (position) controller}
\label{sec_ProposedControl}

The probe dynamics in \eqref{eqn_ConstraintsHAAR} can be written explicitly, using the angular and translational dynamics of the helicopter which results in 
\begin{align} \label{eq_probeDynamics}
   \begin{bmatrix}
       \dot{X}_P \\
       \dot{Y}_P \\
       \dot{Z}_P
   \end{bmatrix}  = 
   \begin{bmatrix}
       \dot{X} \\
       \dot{Y} \\
       \dot{Z}
   \end{bmatrix} + \dot{\mathbf{R}}(\Psi) \bar{x}
\end{align}

where $\Psi = \left[\phi ~\theta ~\psi\right]^T$ defines the attitudes and $\mathbf{R}$ is the transformation matrix from the helicopter to the NED coordinate frame. $\bar{x}$ is the relative distance between the center of gravity of the helicopter and the probe tip in the helicopter coordinate frame. $\dot{X}_P$, $\dot{Y}_P$, $\dot{Z}_P$ are the probe velocities expressed in NED coordinate frame. The goal of the proposed controller is to take the probe position and velocity into account to calculate the commanded accelerations of the CG of the helicopter. Explicitly, they take the closure and its rate in calculating the commanded acceleration. In the proposed controller, we modify the controller given in \eqref{eq_CommAcc} as 
\begin{align} \label{eq_CommAcc_proposed}
    \small
    \begin{bmatrix}
        \ddot{X}_{c} \\
        \ddot{Y}_{c} \\
        \ddot{Z}_{c}
    \end{bmatrix} = 
    \begin{bmatrix}
        \ddot{X}^* \\
        \ddot{Y}^* \\
        \ddot{Z}^* 
    \end{bmatrix} + 
    \begin{bmatrix}
        K_{dX} & 0 & 0\\
        0 & K_{dY} & 0 \\
        0 & 0 & K_{dZ} 
    \end{bmatrix} 
    \left(
    \begin{bmatrix}
        \dot{X}^*_P -  \dot{X}_{D}^* \\
        \dot{Y}^*_P -  \dot{Y}_{D}^* \\
        \dot{Z}^*_P -  \dot{Z}_{D}^* 
    \end{bmatrix} - 
    \begin{bmatrix}
        \dot{X}_P -  \dot{X}_{D} \\
        \dot{Y}_P -  \dot{Y}_{D} \\
        \dot{Z}_P -  \dot{Z}_{D}
    \end{bmatrix} 
    \right) \nonumber \\ 
    + 
    \begin{bmatrix}
        K_{pX} & 0 & 0\\
        0 & K_{pY} & 0 \\
        0 & 0 & K_{pZ} \\
    \end{bmatrix}
        \left(
    \begin{bmatrix}
        X^*_P - X^*_{D} \\
        Y^*_P - Y^*_{D} \\
        Z^*_P - Z^*_{D}
    \end{bmatrix} - 
    \begin{bmatrix}
        X_P - X_{D} \\
        Y_P - Y_{D} \\
        Z_P - Z_{D}
    \end{bmatrix}  
    \right)
\end{align}

where $\dot{X}_P^*$, $\dot{Y}_P^*$, $\dot{Z}_P^*$ correspond to the desired probe position in the NED coordinate frame.

Note that in equation \eqref{eq_CommAcc_proposed}, the commanded acceleration of the helicopter CG is based on the closure and the rate of closure.  

\subsection{Docking error dynamics}

 We assume that the inversion is perfect (i.e., no modeling uncertainty and sufficiently fast inner-loop dynamics), which results in 
\begin{equation}
\label{eq_Hacceleration}
\begin{bmatrix}
    \ddot{X} \\
    \ddot{Y} \\
    \ddot{Z}
\end{bmatrix} = 
\begin{bmatrix}
    \ddot{X}_c \\
    \ddot{Y}_c \\
    \ddot{Z}_c
\end{bmatrix}.
\end{equation}
We define the error of interest (to be used for feedback) as
\begin{equation}
\label{eq_error}
    e = \underbrace{(x^*_P - x^*_{D})}_{\text{desired closure}} -  \underbrace{(x_P - x_{D})}_{\text{actual closure}}
\end{equation}

where 
$   x_P^* = \begin{bmatrix}
       {X}_P^* \\
       {Y}_P^* \\
       {Z}_P^*
   \end{bmatrix} $,
   $   x_P = \begin{bmatrix}
       {X}_P \\
       {Y}_P \\
       {Z}_P
   \end{bmatrix} $,
   $   x_D^* = \begin{bmatrix}
       {X}_D^* \\
       {Y}_D^* \\
       {Z}_D^*
   \end{bmatrix} $,
   $   x_D = \begin{bmatrix}
       {X}_D \\
       {Y}_D \\
       {Z}_D
   \end{bmatrix} $. 
This error incorporates the desired closure $x^*_P - x^*_{D}$ and the actual closure $x_P - x_{D}$ between the probe and the drogue. Then, substituting from \eqref{eq_CommAcc_proposed}, \eqref{eq_Hacceleration} to \eqref{eq_error} we obtain
\begin{equation}
    \label{eq_errorDynamics}
    \ddot{e} = -\mathbf{K_{D}}\dot{e} -\mathbf{K_{P}}e + \ddot{x}_D - \ddot{x}_D^* + \left(\mathbf{\ddot{R}}^*(\Psi) - \mathbf{\ddot{R}}(\Psi)\right)\bar{x}.
\end{equation}
$\mathbf{K_P} = \begin{bmatrix}
        K_{pX} & 0 & 0\\
        0 & K_{pY} & 0\\
        0 & 0 & K_{pZ}
\end{bmatrix}$, 
$\mathbf{K_D} = \begin{bmatrix}
        K_{dX} & 0 & 0\\
        0 & K_{dY} & 0\\
        0 & 0 & K_{dZ}
\end{bmatrix}$,\\
$\ddot{x}_D - \ddot{x}_D^*$ and $\left(\mathbf{\ddot{R}}^*(\Psi) - \mathbf{\ddot{R}}(\Psi)\right)\bar{x}$  correspond to P gain matrix, D gain matrix, the drogue uncertainty and probe acceleration uncertainty, respectively.        



\section{Stability and docking performance guarantees}
\label{sec_stability}

We now present analytical guarantees for the stability and docking performance of the proposed controller presented in Section \ref{sec_ProposedControl}. It is key to note here that the goal of the proposed controller is to guarantee the docking of the probe into the drogue. Therefore, we provide a guarantee for all error trajectories into a set whose size is dictated by (1) the outer loop controller gains, (2) uncertainty in drogue motion, and (3) uncertainty in acceleration of the probe (because of helicopter dynamics). 

Towards this, we first construct a candidate Lyapunov function of the form $E^T \mathbb{Q} E$ with $E = \begin{bmatrix}
    e \\
    \dot{e}
\end{bmatrix}$
\begin{equation}\label{eqn_Lyap}
    \begin{split}
        V(E) &= \frac{1}{2}
        \begin{bmatrix}e^T & \dot{e}^T \end{bmatrix}
\underbrace{\begin{bmatrix}\mathbf{Q_1}&\mathbf{Q_3}^T\\ \mathbf{Q_3}&\mathbf{Q_4}\end{bmatrix}}_{\mathbf{\mathbb{Q}}}\begin{bmatrix}e\\\dot{e}\end{bmatrix}\\
        &= \frac{1}{2}e^T\mathbf{Q_1}e+\frac{1}{2}\dot{e}^T\mathbf{Q_4}\dot{e}+e^T\mathbf{Q_3}\dot{e}.
    \end{split}
\end{equation}

Note that $\mathbf{Q}_i$, $i={1,3,4}$ are matrices to be designed, where $\mathbf{Q_1}$ is a symmetric positive definite $m\times m$ matrix, $\mathbf{Q_4}$ is a symmetric, positive definite $n\times n$ matrix, and $\mathbf{Q_3}$ is an $m\times n$ matrix.




Taking the time derivative of the Lyapunov function,

\begin{equation} \label{eqn_Vdot}
    \dot{V}(E) = e^T \mathbf{Q_1} \dot{e} + \dot{e}\mathbf{Q_4} \ddot{e} + e^T\mathbf{Q_3} \ddot{e} + \dot{e}^T\mathbf{Q_3}\dot{e}.
\end{equation}

\noindent Using \eqref{eq_errorDynamics}, we have
\begin{multline} 
    \label{eq_Vdot}
    \dot{V}(E) = e^T(\mathbf{Q_1} - \mathbf{Q_4}\mathbf{K_P} - \mathbf{Q_3}\mathbf{K_D})\dot{e}  + \dot{e}^T(-\mathbf{Q_4}\mathbf{K_D} + \mathbf{Q_3})\dot{e} \\
    - e^T \mathbf{Q_3}\mathbf{K_P}e + \dot{e}^T\left(\mathbf{Q_4}(\ddot{x}_D - \ddot{x}_D^*)  +  \mathbf{Q_4} \left(\mathbf{\ddot{R}}(\Psi) - \mathbf{\ddot{R}}^*(\Psi)\right)\bar{x} \right)\\
    + e^T \left( \mathbf{Q_3}(\ddot{x}_D - \ddot{x}_D^*) + \mathbf{Q_3} \left(\mathbf{\ddot{R}}^*(\Psi) - \mathbf{\ddot{R}}(\Psi)\right)\bar{x} \right). 
\end{multline}

\textit{Remark}: For the remainder of this paper, $\| \cdot \|$ refers to the standard vector 2-norm, unless otherwise noted.

\noindent
\textbf{Claim [Ultimate boundedness of error]:}
\textit{Given the error dynamics in \eqref{eq_errorDynamics} with the control law defined in \eqref{eq_CommAcc_proposed}, uncertainty in probe acceleration  bounded 
 by $\|\left(\mathbf{\ddot{R}}^*(\Psi) - \mathbf{\ddot{R}}(\Psi)\right)\bar{x}\| < \delta _R$ and drogue acceleration uncertainty bounded by $\| \ddot{x}_D - \ddot{x}_D^*\| < \delta _D $; if $\mathbf{K_{P}},~\mathbf{K_{D}}\succ 0$, then all error trajectories  $E (\cdot)= \begin{bmatrix}
    e (\cdot)\\
    \dot{e}(\cdot)
\end{bmatrix} ~\in \mathbb{R}^6$}
\textit{are ultimately driven to the invariant set}
\begin{align} \label{eq_invariantSet}
    \Lambda = \left\{ E \; \middle| \; V(E) <  \dfrac{( \delta_D + \delta_R) ^2}{2}  \left( \dfrac{\overline{\sigma}(\mathbf{K_P})}{\underline{\sigma}(\mathbf{K_P})^2} + \dfrac{1}{\underline{\sigma}(\mathbf{K_D})^2} \right) \right\}
\end{align}
\textit{in a finite time $T$ and remain in $\Lambda$ for all time $t > T$.}

\begin{proof}
Since $\mathbf{K_{P},~K_{D}}\succ 0$, there exists an $\epsilon > 0$ such that $\mathbf{K_{P}}\succ 0$, $\mathbf{K_{D}}\succ \epsilon\mathbf{I}$. Choosing  $\mathbf{Q_1}=\mathbf{K_{P}+\epsilon K_{D}}$, $\mathbf{Q_3} = \epsilon \mathbf{I}$ and $\mathbf{Q_4} = \mathbf{I}$, we have that $\mathbb{Q} \succ 0$, by Schur's Lemma. With this choice of $\mathbb{Q}$, the Lyapunov function $V(E)$ is positive definite and radially unbounded.

Using \eqref{eq_Vdot} with the chosen $\mathbb{Q}$ above, 
\begin{multline} 
    \label{eq_Vdot1}
    \dot{V}(E) =  \dot{e}^T(-\mathbf{K_D} + \epsilon \mathbf{I})\dot{e} \\
    - \epsilon e^T \mathbf{K_P}e + \dot{e}^T\left((\ddot{x}_D - \ddot{x}_D^*)  +   \left(\mathbf{\ddot{R}}^*(\Psi) - \mathbf{\ddot{R}}(\Psi)\right)\bar{x} \right)\\
    + \epsilon e^T \left( (\ddot{x}_D - \ddot{x}_D^*) +  \left(\mathbf{\ddot{R}}^*(\Psi) - \mathbf{\ddot{R}}(\Psi)\right)\bar{x} \right).
\end{multline}
Therefore, we have that
\begin{multline} 
    \label{eq_Vdot2}
    \dot{V}(E) \leq  -(\underline{\sigma}(\mathbf{K_D})  - \epsilon) \| \dot{e}\|^2 -  \epsilon \underline{\sigma}(\mathbf{K_P})  \|e \|^2 \\
    + \left( \|\ddot{x}_D - \ddot{x}_D^* \| + \|\left(\mathbf{\ddot{R}}^*(\Psi) - \mathbf{\ddot{R}}(\Psi)\right)\bar{x}\|\right) \|\dot{e}\|\\
    + \epsilon  \left(   \|\ddot{x}_D - \ddot{x}_D^* \| + \|\left(\mathbf{\ddot{R}}^*(\Psi) - \mathbf{\ddot{R}}(\Psi)\right)\bar{x}\| \right) \|e \|.
\end{multline}
Substituting the bounds for the uncertainty in drogue ($\delta_D$) and probe acceleration ($\delta_R$),
\begin{multline} 
    \label{eq_Vdot2}
    \dot{V}(E) \leq  -(\underline{\sigma}(\mathbf{K_D})  - \epsilon) \| \dot{e}\|^2 -  \epsilon \underline{\sigma}(\mathbf{K_P})  \|e \|^2 + \left(\delta_D + \delta_R \right) \|\dot{e}\|\\
    + \epsilon  \left(  \delta_D + \delta_R \right) \|e \|.
\end{multline}

Next, we note that if $E \in \Lambda^c$, then  $\|e \| > \dfrac{(\delta_D + \delta_R)}{\underline{\sigma}(\mathbf{K_P})}$ and $\|\dot{e} \| > \dfrac{(\delta_D + \delta_R)}{\underline{\sigma}(\mathbf{K_D})}$ (We omit this derivation here for brevity).  

Using this with \eqref{eq_Vdot2}, since $\epsilon$ is arbitrarily small and positive, $\dot{V}(E) < 0 \ \ \forall E \in \Lambda^c$. 

Therefore, invoking the ultimate boundedness  theorem (4.18 in \cite{Khalil:1173048}) as $V(E)$ is positive definite, radially unbounded and $\dot{V}(E) < 0 \ \  \forall E \in \Lambda^c$, any trajectory  of $E(\cdot)$ converges to the invariant set
\begin{align*}
    \Lambda = \left\{ E \; \middle| \; V(E) \leq   \dfrac{( \delta_D + \delta_R )^2}{2}   \left( \dfrac{\overline{\sigma}(\mathbf{K_P})}{\underline{\sigma}(\mathbf{K_P})^2} + \dfrac{1}{\underline{\sigma}(\mathbf{K_D})^2} \right) \right\}.
\end{align*}

\end{proof}

Therefore, we can conclude that given the uncertainty bounds $\delta_R$ and $\delta_D$ and outer loop controller gains, the proposed controller drives the docking error trajectories to an invariant set, which characterizes a performance guarantee for completing the docking maneuver successfully. 





\setcounter{theorem}{0}
\begin{corollary} \label{corollary}
Given, $\mathbf{K_{P}},~\mathbf{K_{D}}\succ 0$ and $\delta_D$, $\delta_R >0$, if $E(t_0) \in \Lambda$ ($\Lambda$ as described in \eqref{eq_invariantSet}), $E(t) \in \Lambda$ for all $t > t_0$.
\end{corollary}

\begin{figure}[h!]
\captionsetup{font=footnotesize}
\centering
\vspace{3mm}
\includegraphics[width=0.42\textwidth]{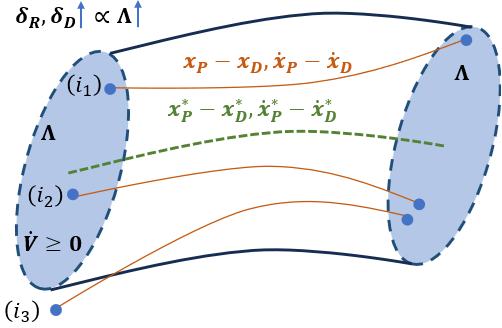}
\caption{Schematic of the invariant set $\Lambda$ for the docking error. Different error trajectories initialized inside the invariant set denoted as $i_{1,2}$ remain in the invariant set. Error trajectory $i_{3}$ initialized outside the invariant set, converges to this invariant set. Increasing the uncertainty terms $\delta_D$, $\delta_R$ expands $\Lambda$ whereas the decreasing results in a smaller set $\Lambda$.}
\label{Fig_LyapSet}
\end{figure}

\noindent \textbf{Interpretation of the guarantee assured by the invariant set}: The result given in \textbf{Corollary 1} guarantees that the docking error trajectories remain within an invariant set when the drogue uncertainty and probe acceleration uncertainty are both bounded. Increasing the outer loop controller gains shrinks the size of this set while increasing uncertainty in drogue motion and angular acceleration of the helicopter results in inflating the size of the set, which is to be expected. It should be noted that setting excessively high controller gains will also result in a higher acceleration of the helicopter CG (refer \eqref{eq_CommAcc_proposed}), resulting in actuator saturation and a larger residue of angular acceleration $\delta_R$.  Figure \ref{Fig_LyapSet} illustrates a visual representation of the invariant set $\Lambda$. 

\section{Simulation setup}
\label{sec_simSetup}

We now describe the high-fidelity simulation set-up used to validate the proposed controller and ultimate boundedness guarantees. 

\subsection{Helicopter dynamic model} \label{sec_FSHCModel}

For validation, we use a high-fidelity helicopter simulation platform \cite{Krishnamurthi_2017_Flight}, based on a UH60 Black Hawk model \cite{Howlett_1981_UH}. The dynamics, in general, are given by $\dot{\mathbf{x}}=\mathbf{f}(\mathbf{x},\mathbf{u})$ where $\mathbf{x}$ is the state vector $\mathbf{x}=\begin{bmatrix}
\mathbf{x}_f^T,\mathbf{x}_r^T,\mathbf{x}_t^T,\mathbf{x}_e^T
\end{bmatrix}^T$ where 
$\mathbf{x}_f=\begin{bmatrix}
u,v,w,p,q,r,\phi,\theta,\psi,X,Y,Z
\end{bmatrix}^T$, 
$\mathbf{x}_r=\begin{bmatrix}
\beta_0,\beta_{1s},\beta_{1c},\beta_{d},\dot{\beta}_0,\dot{\beta}_{1s},\dot{\beta}_{1c},\dot{\beta}_{d},\lambda_0,\lambda_{1s},\lambda_{1c}
\end{bmatrix}^T$, 
$\mathbf{x}_t=\lambda_{0TR}$, 
$\mathbf{x}_e=\begin{bmatrix}
\Omega,\chi_f,Q_e
\end{bmatrix}^T$. $\mathbf{x}_f$ denotes 12 fuselage rigid body states, $\mathbf{x}_r$ denotes 8 blade flapping states and 3 inflow states of the main rotor, $\mathbf{x}_t$ denotes the tail rotor inflow state, $\mathbf{x}_e$ denotes 3 engine states.
\noindent The control input $\mathbf{u}$ is given by $\mathbf{u}=\begin{bmatrix}
u_{lat},u_{long},u_{col},u_{ped},u_{tht}
\end{bmatrix}^T$ which consists of lateral, longitudinal, and collective joystick input to the main rotor, pedal input to the tail rotor, and throttle input to the engine. Note that fuselage input $\mathbf{u}_f=\begin{bmatrix}
u_{lat},u_{long},u_{col},u_{ped}
\end{bmatrix}^T$ is comprised of the input channels governing the fuselage motion.

\subsection{Drogue dynamics} \label{sec_DrogueModel}

For drogue dynamics, we use the high-fidelity drogue model developed in \cite{boothe2018introduction} to generate drogue state trajectories. This model represents the hose-drogue as a set of links that are connected in series, and accurately models the aerodynamic, torsional, bending, and gravitational forces acting on the hose and the drag and lift forces acting on the drogue. These forces translate into accelerations that result in time-varying positions and velocities in the inertial frame denoted as ${x}_{D} \triangleq [X_D, Y_D, Z_D, u_D, v_D, w_D]$. In particular, we generate varying ${x}_{D}$ trajectories with varying wind conditions.

\section{Computational experiments}

In this section, we illustrate how the proposed controller is implemented in an autonomous helicopter aerial refueling maneuver. We employ the simulation setup described in Section    \ref{sec_simSetup}. Then, we visualize how the error state evolves with time which ultimately converges to the invariant set described in Section \ref{sec_stability}. 

\subsection{Performance evaluation of the proposed controller}

In this section, we evaluate the performance of the proposed controller in Section \ref{sec_ProposedControl} and compare it to the standard position controller (refer to Section \ref{sec_StandardControl}) which does not take the orientation of the helicopter into account. 

At the trajectory planning stage, we assume wind-free conditions, i.e., in $x_D^* = [X_{D0} + v_T \cdot t, ~Y_{D0}, ~Z_{D0}, ~v_T, ~0,~0]$ where $X_{D0}$, $Y_{D0} = 0m$, $Z_{D0} = -1000m$ and $v_T = 56.58 m/s$.  At the start of the maneuver, the initial positions of the helicopter CG and the drogue are given by ($0+\delta_0$, $0+\delta_0$, $-1000+\delta_0$)$m$ and (5, 0, -1000)$m$, respectively in the NED coordinate frame where $\delta_0 \sim U \left[ -0.2,0.2 \right]m$. Furthermore, different drogue trajectories are simulated by introducing a crosswind of which the magnitude is varied uniformly as $~U[-5, 5] kt$ in the horizontal plane. The PD gains of the proposed controllers were chosen as

$\mathbf{K_P} = \begin{bmatrix}
        0.41 & 0 & 0\\
        0 & 0.37 & 0\\
        0 & 0 & 35
\end{bmatrix}$, 
$\mathbf{K_D} = \begin{bmatrix}
        0.75 & 0 & 0\\
        0 & 0.75 & 0\\
        0 & 0 & 8.8
\end{bmatrix}$. 

The proposed controller is deployed in the high-fidelity simulation platform described in Section \ref{sec_simSetup}.

\begin{figure}[h!]
\captionsetup{font=footnotesize}
\centering
\vspace{2mm}
\includegraphics[trim={1mm 0mm 0mm 0mm}, clip, width=0.49\textwidth]{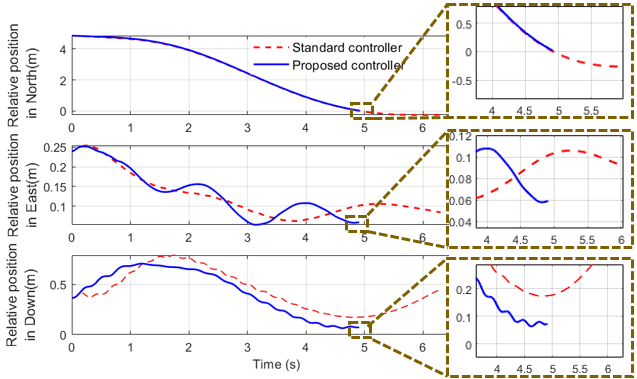}
\caption{Comparison between the proposed controller and the standard controller.}
\label{Fig_dockingComp}
\end{figure}

Figure \ref{Fig_dockingComp} illustrates the sample trajectories of the error between the drogue and the probe for 1) the standard controller and 2) the proposed controller. Note that the trajectories corresponding to the proposed controller result in a docking error  $\| x_P -x_D \| \leq 0.2m$ deemed a successful docking. Then, we extend the experiment to 50 simulations with different initial helicopter positions (CG locations) and randomized wind conditions that impact the drogue movement. The results obtained in these 50 simulations are given in Table \ref{Tab_Performance}.

\begin{table}[h!]
\centering
\caption{Results from 50 simulated docking maneuvers}
\label{Tab_Performance}
\begin{tabular}{|l|l|l|}
\hline
Criterion & Standard  & Proposed \\
& controller & controller\\ \hline 
Mean docking error(m)   & 0.22  & 0.14 \\ 
\hline    
Docking error St. dev. (m)  & 0.09  & 0.07 \\ \hline
Docking success rate & 38\%  & 74\% \\ \hline
\end{tabular}
\end{table}

We observe that we achieve a docking success of 74\%, with a mean docking error of 0.14m (std. dev.: 0.07m) when the proposed controller is employed. In contrast, when the standard control structure is employed, we achieve 38\% docking and a mean docking error of 0.22m (std. dev.: 0.09m). Thus, the proposed controller shows a 36\% improvement over the standard controller in the 2-norm docking error, achieving a docking success rate that is double that of the standard approach.

\begin{figure}[h!]
\captionsetup{font=footnotesize}
\centering
\includegraphics[trim={8mm 0mm 8mm 4mm}, clip, width=0.47\textwidth]{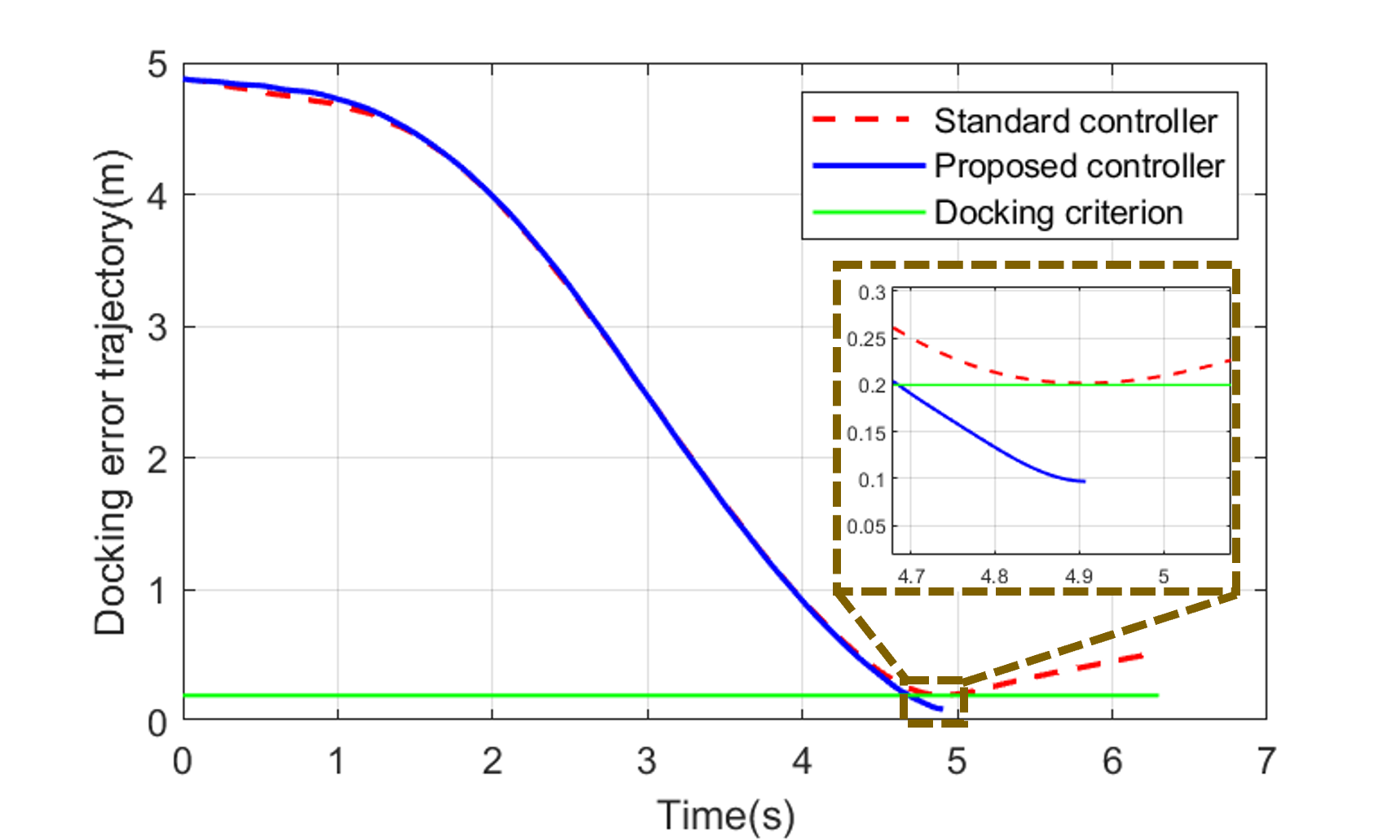}
\caption{2-norm docking error at the time of docking when the proposed controller is utilized.}
\label{Fig_dockingError}
\end{figure}

Figure \ref{Fig_dockingError} illustrates a sample docking trajectory $\| x_P (t) - x_D (t)\|$. Note that the docking error under the proposed controller goes to $0.2m$ whereas the standard controller fails to satisfy the docking criterion.

\begin{figure}[h!]
\captionsetup{font=footnotesize}
\centering
\includegraphics[trim={8mm 1mm 8mm 5mm}, clip, width=0.51\textwidth]{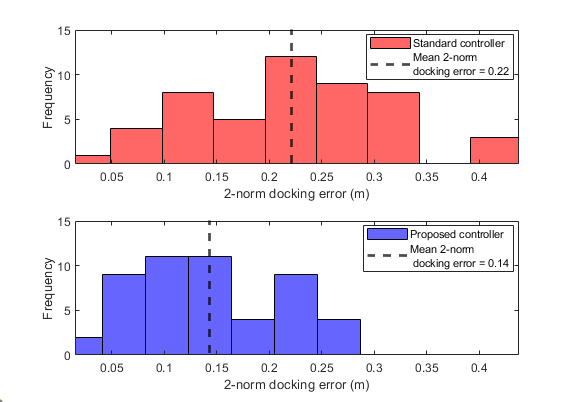}
\caption{Histograms of 2-norm docking error at the time of contact for 50 simulation runs.}
\label{Fig_YZError}
\end{figure}

For the 50 simulation runs, we plot the 2-norm docking error at the time of docking as shown in Figure \ref{Fig_YZError}. We notice that the proposed controller outperforms the standard controller.

\subsection{Comparison of docking performance against theoretical bound}

\begin{figure}[h!]
\captionsetup{font=footnotesize}
\centering
\includegraphics[trim={8mm 0mm 5mm 2mm}, clip, width=0.50\textwidth]{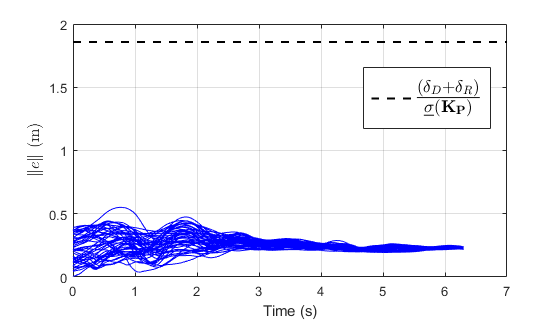}
\caption{Evolution of the 2-norm error ($\|e\|_2$) over 50 simulation runs. The computed upper bound guarantee is shown in the dashed black line.}
\label{Fig_Lyapunov}
\end{figure}

This section discusses the theoretical result derived in Section \ref{sec_stability} about the performance of the proposed controller. Figure \ref{Fig_Lyapunov} shows the evolution of the norm of positional error states ($\| e\|_2$) with time for 50 simulations. In the same plot, we indicate the bound of the 2-norm positional error $\|e \| > \dfrac{(\delta_D + \delta_R)}{\underline{\sigma}(\mathbf{K_P})}$ given in Section \ref{sec_stability}. The bound of the uncertainty in the drogue acceleration  $\delta_D$ is $0.18m/s^2$ and the bound of the uncertainty in the probe acceleration(due to attitude changes) $\delta_D$ is $0.51m/s^2$. As shown in Figure \ref{Fig_Lyapunov}, the positional error trajectories do not exceed the theoretical bound. This guarantees the fact that the 2-norm of error $\| e\|_2$ between the position of the probe and the drogue is always bounded by the 1.8m under bounded uncertainty.  




\section{Conclusions and Future Work} 

In this paper, we presented a new control method that applies to autonomous helicopter aerial refueling. While keeping the foundation on a generic dynamic inversion-based inner-loop outer-loop control architecture, we designed a controller that takes the orientation of the probe into account. Then, we proved the stability of the proposed controller using Lyapunov analysis. Furthermore, we quantified the performance of the proposed controller employing the theory of ultimate boundedness under uncertainty due to the drogue movement and attitude acceleration. Finally, we validated the proposed controller in a high-fidelity helicopter simulation platform with a realistic UH60 Black Hawk model and a dynamic drogue model. The conducted numerical experiments showcase the effectiveness of the proposed controller. The performance can be improved by incorporating real-time trajectory generation capabilities with improved disturbance-rejecting properties. 

\section*{Acknowledgement}
This work was sponsored by the Office of Naval Research (ONR), under contract number N00014-23-1-2377.

\bibliographystyle{IEEEtran} 
\bibliography{main}




\end{document}